%% file: root.tex
\documentclass[conference]{IEEEtran}
\usepackage[letterpaper, margin=0.75in]{geometry}
\usepackage{epsfig}
\usepackage{amsmath}
\usepackage{amssymb}
\usepackage{float}
\usepackage{subcaption}  
\usepackage{afterpage}

\usepackage[breaklinks=true,colorlinks,bookmarks=false]{hyperref}

\newcommand{\etal}{\textit{et al.} }

\ifCLASSINFOpdf
\else
\fi
\begin{document}
%



\title{
Instance Segmentation as\\
Image Segmentation Annotation
}

\author{
  \IEEEauthorblockN{Thomio Watanabe}
  \IEEEauthorblockA{University of Sao Paulo\\
  {\tt\small thomio.watanabe@usp.br}}
  \and
  \IEEEauthorblockN{Denis Wolf}
  \IEEEauthorblockA{University of Sao Paulo\\
  {\tt\small denis@icmc.usp.br}}
}

\newgeometry{top=1in,bottom=0.75in,right=0.75in,left=0.75in}
\maketitle



%

\input{src/abstract.tex}
\input{src/intro.tex}

\afterpage{\aftergroup\restoregeometry}
\input{src/related_work.tex}

\input{src/instance_classification.tex}

\input{src/loss_function.tex}

\input{src/experiments.tex}

\input{src/conclusion.tex}
\input{src/acknowledgment.tex}

\bibliographystyle{ai_bibtex/IEEEtran}
\bibliography{ai_bibtex/references}


%
%

\end{document}

%% file: src/abstract.tex
\begin{abstract}
The instance segmentation problem intends to precisely detect and delineate objects in images.
Most of the current solutions rely on deep convolutional neural networks but despite this fact proposed solutions are very diverse.
Some solutions approach the problem as a network problem, where they use several networks or specialize a single network to solve several tasks.
A different approach tries to solve the problem as an annotation problem, where the instance information is encoded in a mathematical representation.
This work proposes a solution based in the DCME technique to solve the instance segmentation with a single segmentation network.
Different from others, the segmentation network decoder is not specialized in a multi-task network.
Instead, the network encoder is repurposed to classify image objects, reducing the computational cost of the solution.
\end{abstract}

%% file: src/intro.tex
\section{Introduction}
\label{sec:intro}

Most instance segmentation solutions are based in a combination of multiple convolutional neural networks.
These solutions present high scores on benchmark tests at the price of a high computational cost.
They usually separate the solution in subtasks and they try to solve them separately.
Although this approach is more clear and easy to explain, it may be too restrictive to neural networks.
Once neural networks use examples to find the best features to solve a problem, directly exposing the problem to a network may provide more efficient solutions.
This work investigates this hypothesis and proposes a solution for the instance segmentation problem. 

As proposed by \text{Watanabe and Wolf} \cite{watanabe2018dcme}, the DCME representation and was able to generate class-agnostic instances masks.
We extend this previous work using a single encoder-decoder CNN for image segmentation.
We repurpose the network encoder to remove the classification network from the solution pipeline.
Also we modify the network decoder regression function to become more robust to large input image resolutions.
For the best of our knowledge, this is the only solution that solves the instance segmentation problem with a single segmentation network without specializing the network decoder in multiple branches (multi-task networks).

This approach changes the focus of most researches, where different network architectures are proposed to solve the problem.
Instance segmentation scores will improve with better segmentation models.
This approach adds overhead to decode instance masks but its computational cost is smaller than adding a neural network to solve a subtask.

Usually, a segmentation network encoder output is never directly used as part of an instance segmentation solution.
In this work we repurpose the network encoder to classify instance masks.
This approach is more computational efficient when compared to solutions that specialize a decoder branch in a segmentation model.
We also assume that repurposing the network encoder is more suitable for the instance segmentation problem.
The classification task have a strong prior where the input image has only a single centralized object and this assumption may hinder the decoder optimization.

During the experiments, it was observed that the DCME was very susceptible to large input image resolutions.
This behavior was recognized as the high output loss signal due the long distances vectors in the objects borders.
To improve results two main modifications were made in the loss function.
We consider each pixel from each output channel as an independent output and we clip the loss error before updating the backpropagation.



This work uses deep Convolutional Neural Networks (CNNs) for image segmentation \cite{garcia2017review} to learn the DCME technique.
Specifically, the segmentation model was based in the \text{DeepLabv3+} network proposed by Chen \etal \cite{chen2018encoder}.
The proposed solution was evaluated in the Cityscapes \cite{Cordts2016Cityscapes} dataset, an urban roads scenes dataset.

%% file: src/related_work.tex
\section{Related Work}

Several solutions for the instance segmentation problem are based in detection networks where they try to extract object masks from object detections.
Solutions based in object detections are being denominated proposal-based networks and they propose different network architectures to improve results.
This is the main approach for instance segmentation solutions which are able to achieve high benchmark scores.
However these solutions have a high computational cost associated with the extensive use of convolutional layers.

For instance, the Mask R-CNN \cite{he2017mask} extends the Faster R-CNN \cite{ren2015faster} network by predicting masks in parallel with object bounding boxes.
Other recent and notable work is the Path Aggregation Network \cite{liu2018path} which shortens the information path between lower and top layers.

In opposite, solutions that do not rely on object detections are being denominated proposal-free networks.
These solutions are diverse and present different encoding techniques, clustering techniques, loss functions and multi-task networks \cite{uhrig2016pixel, de2017semantic, bai2017deep, liu2017sgn, kendall2018multi}.
For a broader review, \text{Watanabe and Wolf} \cite{watanabe2018dcme} summarizes recent researches. 

As previously stated, this work extends the DCME \cite{watanabe2018dcme} removing the classification network and improving the loss (objective) function.
Concurrently with \text{Watanabe and Wolf} \cite{watanabe2018dcme}, Kendall \etal \cite{kendall2018multi} also presented a similar encoding technique.
However, Kendall \etal \cite{kendall2018multi} proposed a multi-task network to generate vectors maps, segmentation maps and depth maps.
And, to optimize the multi-task network they have proposed a loss function to learn coefficients to ponder each task loss.

%% file: src/instance_classification.tex
\section{Instance classification}
Most of the CNNs for image segmentation have an encoder-decoder architecture.
These networks present different decoders but usually they reuse the encoder from classification networks like VGG \cite{simonyan2014very} or ResNet\cite{he2016deep}.
However, reusing classification networks might be disadvantageous for the instance segmentation problem.

In the classification problem there is only a single object in the image and it is usually centralized.
Therefore, these networks do not need to learn the number of objects and their positions in the image.
For the detection problem there is a variable number of objects from different classes and they can be located anywhere.

If the image has several objects it makes more sense to classify parts of it.
Based on this assumption we decided to repurpose the encoder to perform a segmentation.
Although it is similar to the image segmentation, its main purpose is to roughly localize and classify the objects in the image.

The input image spatial dimensions are reduced in every stride operation higher than one.
In the most common classification networks the stride is equal to 2 in horizontal and vertical directions.
We define the grid size, $G_s$, as the final encoder division number, from the input image to the encoder output.
For $n$ $(2,2)$ stride operations the grid size is given by Equation \ref{eq:grid_size}, and it is the same for both horizontal and vertical dimensions.
\begin{equation}
  G_s = 2^n
  \label{eq:grid_size}
\end{equation}

\begin{figure}[ht]
  \centering
  \includegraphics[width=0.8\linewidth]{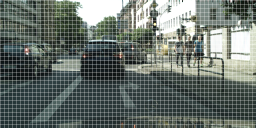}
  \vspace{1mm}\\
  \includegraphics[width=0.8\linewidth]{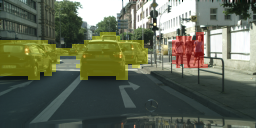}
  \caption{ \textbf{Top:} input image with (512,1024) resolution and grid size equals to 16. Presents 32 vertical blocks and 64 horizontal blocks. \textbf{Bottom:} Input image blocks classification. Car class in yellow, people class in red. }
  \label{fig:grid}
\end{figure}

The grid size defines the size of the input image grid as depicted in Figure \ref{fig:grid}.
Each encoder output infers the class of one grid block.
During the training process, class labels are defined according to Equation \ref{eq:image_to_encoder},
 where $P(x,y)_{image}$ is the image pixel position and $P(x,y)_{encoder}$ is the encoder output position.

\begin{equation}
  \begin{aligned}
    P\left(x,y\right)_{encoder} &= \textit{floor} \left( \frac{ P\left(x,y\right)_{image} }{ G_s } \right)
  \end{aligned}
  \label{eq:image_to_encoder}
\end{equation}

When generating the annotations, a grid block may have pixels from more than one class and to solve this problem we define a priority list.
Underrepresented classes with smaller number of instances have the preference to label the block.

The Equation \ref{eq:encoder_to_image} associates the encoder output position to each input image block
 where $ P\left(x_{0},y_{0}\right)_{image} $ is the top left point from each grid block.
The bottom right point from the block is given by adding the grid size in both dimensions.
All pixels from the top left point to the bottom right point are labeled with the same class.
It is an open interval and the higher limits are not included.

\begin{equation}
  \begin{aligned}
    P\left(x_{0},y_{0}\right)_{image} &= G_s \cdot P\left(x,y\right)_{encoder} \\
  \end{aligned}
  \label{eq:encoder_to_image}
\end{equation}


Since our main purpose is to solve the instance segmentation there is no need to infer the class of each input block, like in Figure \ref{fig:grid}.
To find out the class of the instances we only use Equation \ref{eq:image_to_encoder} to infer the class of the block that contains the DCME instance center of mass.

%% file: src/loss_function.tex
\section{Network decoder loss function}
In the image classification problem the exact position and size of the objects are not important.
Therefore, to reduce computational costs the image resolution is usually reduced to $(227,227)$ and classification networks are built taking this in account. 
For image segmentation and object detection high resolution images are important to get precise results.
Since larger images highly increase the computational cost, a resolution/precision trade-off becomes an intrinsic part of the solution.

When compared to multi-tasks networks, reusing the network encoder for classifying instances reduces the computational cost of the solution.
And, given a fixed resource capabilities, this allow us to increase the input image resolution.

\begin{figure}[ht]
  \centering
  \includegraphics[width=0.8\linewidth]{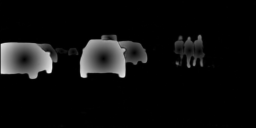}
  \caption{ DCME magnitude map. Large objects present longer vectors close to their borders. Car borders are brighter than people borders. }
  \label{fig:magnitude}
\end{figure}

During experiments, it was observed that deep CNNs were not able to learn and generalize the DCME encoding for large image resolutions.
Once the DCME encoding is based on 2D displacement vectors, changing the image resolution directly affect vectors sizes.
Small image resolutions will reduce the presence of small objects.
In opposite, large resolutions present too large objects that 
not only have more vectors but also have very long vectors close to their borders, Figure \ref{fig:magnitude}.

Errors in bigger objects will generate very high error values while errors in small objects will be insignificant.
This behavior generate biased models that will preferably detect large objects.
To make the DCME more robust to object sizes, two main modification were proposed in the decoder loss function.

\subsection{Independent outputs}

\text{Watanabe and Wolf} \cite{watanabe2018dcme} used the Caffe \cite{jia2014caffe} Euclidean loss function to compute the DCME regression.
This function calculates the loss according to the Mean Squared Error (MSE), Equation \ref{eq:mse}.
Where $N$ is the number of samples, $Y_i$ is a label and $\hat{Y}_i$ the model output.
Furthermore, the same loss signal was used to update all backpropagation differences in the last layer.
This approach presented a high loss error signal which was compensated by a small learning rate.

\begin{equation}
  \text{MSE} = \frac{1}{2N} \sum^N_{i=1} \left( Y_i - \hat{Y}_i \right)^2
  \label{eq:mse}
\end{equation}

Since each output is independent of each other and they present different values, applying a single mean value to compute the gradients was a major flaw.

To overcome this problem, every single output pixel from both output channels in the network decoder were considered as an independent output.
Each output loss was directly updated with its corresponding error and the model general loss was evaluated considering this, with the number of samples defined by Equation \ref{eq:general_loss}.
The number of samples is given by $2$ DCME output channels, the number of images $n$ and the decoder spatial dimensions $(r,c)$.

\begin{equation}
  N = 2 \cdot n \cdot r \cdot c
  \label{eq:general_loss}
\end{equation}



\subsection{Error amplitude}

Even with this previous modification, the segmentation model was still susceptible to high resolution images (large objects).
To solve this problem the error values were clipped before the backpropagation.
We used a modified version of the logistic function, defined in Equation \ref{eq:loss}.
The function is symmetric to the origin and does not only clip the loss but also presents a linear behavior around the origin.
The function amplitude is ${A}/{2}$.

\begin{equation}
  \begin{aligned}
    f(x) &= A \cdot \left( \frac{1}{1 + e^{-x}} - 0.5 \right)\\
         &= \frac{A}{2} \cdot \left( \frac{e^{x} - 1}{e^{x} + 1} \right)
    \label{eq:loss}
  \end{aligned}
\end{equation}

\begin{figure}[ht]
  \includegraphics[scale=.4]{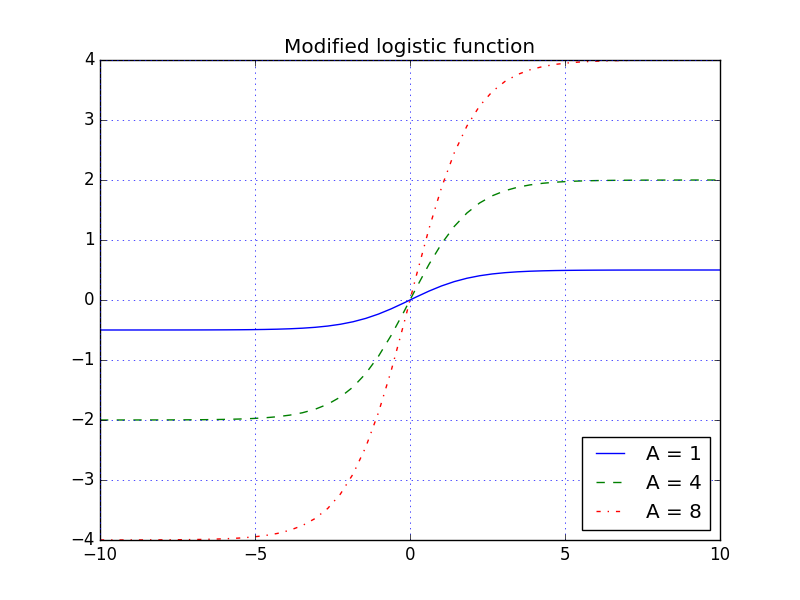}
  \caption{Translated logistic function used to clip error differences. The function is differentiable, is symmetric to the origin and linear around zero.}
  \label{fig:loss}
\end{figure}

In this approach $A$ is a parameter that must be adjusted according to the input image resolution.
The Figure \ref{fig:loss} illustrates the Equation \ref{eq:loss} for different values of $A$.

It is interesting to note that the error values are clipped before the backpropagation.
The loss function output is calculated with the full error values and it gives a realistic estimate of the model learning capabilities.

%% file: src/experiments.tex
\section{Ablation Studies}

The ground truth masks were exactly implemented according to the DCME \cite{watanabe2018dcme} and the experiments were performed on the Cityscapes dataset \cite{Cordts2016Cityscapes}.
The solution was tested in the validation set and, to better evaluate the generalization error, the validation set was never used for training purposes.
All experiments were performed with the image resolution at $(512,1024)$, and a clip amplitude of two, $A = 4$.

The segmentation model was based in the \text{DeepLabv3+} \textit{encoder-decoder with atrous convolution} network proposed by Chen \etal \cite{chen2018encoder}.
Differently from the authors, we used a VGG encoder with its corresponding ImageNet weights.
We have also added 5 convolutional layers to the end of the encoder.
In the decoder, the upsample layers were replaced by deconvolution layers, once they were more stable during training.
The network was implemented in Caffe \cite{jia2014caffe}.

The segmentation model is one of the most important parts of this solution.
The maximum mean average precision (AP) we could obtain with \text{DeepLabv3+} was 11.5 AP.
This solution was tested with SegNet \cite{badrinarayanan2015segnet} and the highest mean AP it could provide was 7.5 AP.
Therefore, only switching segmentation models the scores had a mean increase of 53,33\%.




The \text{DeepLabv3+} was trained with a batch size of 6 images in a single 11GB GPU.
This model is more efficient than the SegNet, where the largest batch size that could fit the GPU was 3 images.
Compared to the DCME \cite{watanabe2018dcme} that uses 2 CNNs, this solution only requires a single segmentation network which is even more computational efficient than the their segmentation network.

\begin{figure*}[ht]
\begin{center}
	  \begin{subfigure}{\textwidth}
        \includegraphics[scale=.65]{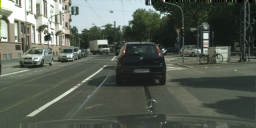}
        \vspace{1mm}
        \includegraphics[scale=.65]{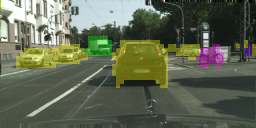}
        \includegraphics[scale=.65]{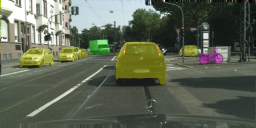}
    \end{subfigure}
    \vspace{1mm}
    \begin{subfigure}{\textwidth}
        \includegraphics[scale=.65]{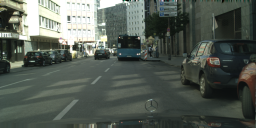}
        \includegraphics[scale=.65]{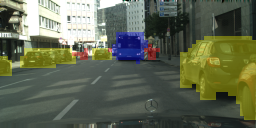}
        \includegraphics[scale=.65]{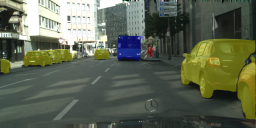}
    \end{subfigure}
    \vspace{1mm}
    \begin{subfigure}{\textwidth}
        \includegraphics[scale=.65]{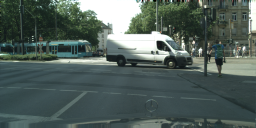}
        \includegraphics[scale=.65]{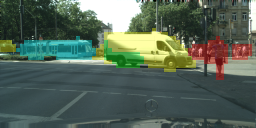}
        \includegraphics[scale=.65]{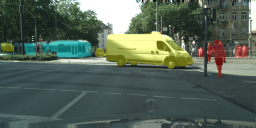}
    \end{subfigure}
    \vspace{1mm}
    \begin{subfigure}{\textwidth}
        \includegraphics[scale=.65]{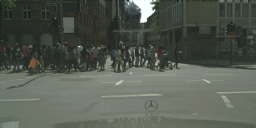}
        \includegraphics[scale=.65]{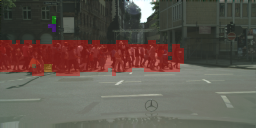}
        \includegraphics[scale=.65]{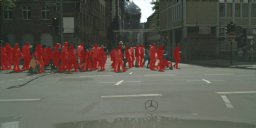}
    \end{subfigure}
    \vspace{1mm}
    \begin{subfigure}{\textwidth}
        \includegraphics[scale=.65]{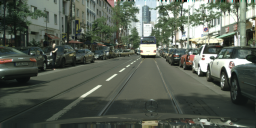}
        \includegraphics[scale=.65]{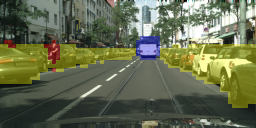}
        \includegraphics[scale=.65]{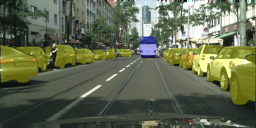}
    \end{subfigure}
    \vspace{1mm}
\end{center}
\caption{\textbf{From left to right:} input images, class map and instance masks on cityscapes validation set.}
\label{fig:validation_images}
\end{figure*}

The best results achieved in the validation set are detailed in Table \ref{tab:validation_results} (left) and Figure \ref{fig:validation_images}.
The table on the right presents results from the validation set ground truth.
Just because the dataset was resized by half in both directions the mean AP was reduced to $59.2\%$.
Once the downsampling operation loses information that is never retrieved, the upsampled masks are less precise than the original ones.

\begin{table}[ht]
\centering
\caption{ \textbf{Left:} best results on validation set. \textbf{Right:} ground truth evaluation for $(512,1024)$ resolution. }
\label{tab:validation_results}
\scalebox{1.}{
\begin{tabular}{ l r r }
  \hline
  Class      &   AP & AP50\% \\
  \hline
  person     &  6.8 & 16.6 \\
  rider      &  4.0 & 12.3 \\
  car        & 24.0 & 37.2 \\
  truck      & 10.9 & 15.3 \\
  bus        & 20.3 & 27.9 \\
  train      & 22.1 & 37.3 \\
  motorcycle &  2.0 &  6.7 \\
  bicycle    &  1.9 &  6.3 \\
  \hline
  mean       & 11.5 & 20.0 \\
  \hline
\end{tabular}

\quad

\begin{tabular}{ l r r }
  \hline
  Class      &   AP & AP50\% \\
  \hline
  person     & 45.7 &  96.3 \\
  rider      & 52.6 &  98.8 \\
  car        & 57.7 &  97.2 \\
  truck      & 66.5 & 100.0 \\
  bus        & 77.8 & 100.0 \\
  train      & 75.7 & 100.0 \\
  motorcycle & 49.5 &  96.0 \\
  bicycle    & 47.7 &  93.9 \\
  \hline
  mean       & 59.2 & 97.8 \\
  \hline
\end{tabular}
}
\end{table}

We have designed two different experiments to separately evaluate the encoder and the decoder performance.
The \textit{instance oracle} experiment evaluates the encoder and assumes all instances are perfectly detected.
The \textit{class oracle} experiment evaluates the decoder and provides the correct class of all detected instances.



\subsection{Instance oracle}

In this experiment the validation set ground truth was used to correctly detect all instances.
A class predicted by the encoder is associated to the instance CM with the Equation \ref{eq:image_to_encoder}.
The mean AP presented by the instance oracle, Table \ref{tab:instance_oracle}, is around 42.57\% of the ground truth, which is a considerably high value.

\begin{table}[ht]
\centering
\caption{ Per class average precision on validation set. Detection with instance oracle and classification with network encoder.  }
\label{tab:instance_oracle}
\scalebox{1.1}{
\begin{tabular}{ l r r }
  \hline
  Class      &   AP & AP50\% \\
  \hline
  person     & 29.0 & 52.2 \\
  rider      & 20.1 & 29.6 \\
  car        & 48.2 & 75.6 \\
  truck      & 20.6 & 22.1 \\
  bus        & 27.8 & 31.9 \\
  train      & 24.8 & 27.3 \\
  motorcycle & 13.7 & 20.0 \\
  bicycle    & 17.7 & 28.1 \\
  \hline
  mean       & 25.2 & 35.9 \\
  \hline
\end{tabular}
}
\end{table}

The object detections were also used to evaluate the encoder classification accuracy.
For each detection we compared its ground truth class and its prediction.
Its global accuracy was $64.97\%$ and a per class evaluation is presented in Table \ref{tab:class_accuracy}.

\begin{table}[ht]
\centering
\caption{ Instance classification with encoder. Number of instances, correct classification and accuracy. }
\label{tab:class_accuracy}
\scalebox{1.1}{
\begin{tabular}{ l r r r }
  \hline
  Class      & Instances &  Correct & Acc.\% \\
  \hline
  person     & 3394 & 1974 & 58.16 \\
  rider      &  543 &  230 & 42.36 \\
  car        & 4653 & 3847 & 82.68 \\
  truck      &   93 &   29 & 31.18 \\
  bus        &   98 &   44 & 44.90 \\
  train      &   23 &   10 & 43.48 \\
  motorcycle &  149 &   41 & 27.52 \\
  bicycle    & 1165 &  399 & 34.25 \\
  \hline
  total      & 10118 & 6574 & 64.97 \\
  \hline
\end{tabular}
}
\end{table}

The classification accuracy was far from ideal.
Classes with small number of samples like truck, bus, train and motorcycle presented a small accuracy which is due to the dataset imbalance.
The bicycle class should have presented higher scores but it is a particularly hard object to detect and classify.


\subsection{Class oracle}
This experiment used the provided class labels to evaluate the masks generated by the decoder.
It evaluates if this solution is able to find the objects and also if it is able to delineate object contours.
The instances were detected and delineated by the DCME and the classes were obtained from the validation set ground truth.
An evaluation is presented for each class in Table \ref{tab:class_oracle}.

\begin{table}[ht]
\centering
\caption{ Per class average precision on validation set. Classification with class oracle and detection with decoder. }
\label{tab:class_oracle}
\scalebox{1.1}{
\begin{tabular}{ l r r }
  \hline
  Class      &   AP & AP50\% \\
  \hline
  person     &  7.2 & 17.8 \\
  rider      &  4.2 & 12.9 \\
  car        & 24.5 & 37.9 \\
  truck      & 14.1 & 23.7 \\
  bus        & 30.1 & 47.1 \\
  train      & 25.6 & 42.7 \\
  motorcycle &  3.8 & 12.4 \\
  bicycle    &  2.5 &  8.5 \\
  \hline
  mean       & 14.0 & 25.4 \\
  \hline
\end{tabular}
}
\end{table}

Table \ref{tab:class_oracle} mean AP represent 23.65\% of the ground truth mean AP, Table \ref{tab:validation_results} (right).
The mean AP difference with the complete solution, Table \ref{tab:validation_results} (left), was only $2.5$.
When compared to the instance oracle this low score indicates the solution bottleneck is the decoder/detection. 

Since the DCME does not distinguish objects in classes, it is more robust against data imbalance.
This is noticeable on scores obtained in the bus and train classes.
The train class have the smallest number of samples but present a score higher than the car class.

An experiment to evaluate the detection accuracy was also elaborated.
A detection was considered as correct if the intersection over union (IoU) of the predicted mask and the ground truth was above $25\%$, $50\%$ or $75\%$ , Table \ref{tab:detection_accuracy}.



\begin{table}[ht]
\centering
\caption{ Detection accuracy for different overlapping thresholds. Total number of instances 10118.  }
\label{tab:detection_accuracy}
\scalebox{1.1}{
\begin{tabular}{ l r r }
  \hline
  threshold(\%)  & detections & Acc(\%) \\
  \hline
  25  & 3821 & 37.76 \\
  50  & 3102 & 30.66 \\
  75  & 1886 & 18.64 \\
  \hline
\end{tabular}
}
\end{table}

In Table \ref{tab:detection_accuracy}, the detection accuracy quickly decreases with higher IoU thresholds.
This means the solution is able to find several objects but it is not able to precisely delineate them.

In Cityscapes, the mean AP metric utilizes multiple IoU thresholds, ranging from 0.5 to 0.95 in steps of 0.05, \cite{hariharan2014simultaneous, lin2014microsoft}.
Higher thresholds require precise masks but the average over multiple thresholds lessens this requirement.

When compared to the instance oracle, it is noticeable that the average precision (AP) metric is more severe with detection errors than with classification errors.
If the solution is not able to detect an object its classification is not even considered by the evaluation metric.

\subsection{Test set evaluation}

The Cityscapes test set was also evaluated and the results were submitted to the online server.
The per class evaluation is presented in Table \ref{tab:test_evaluation}.

\begin{table}[ht]
\centering
\caption{ Per class evaluation on test set. }
\label{tab:test_evaluation}
\scalebox{1.1}{
\begin{tabular}{ l r r }
  \hline
  Class      &    AP & AP50\% \\
  \hline
  person     &  6.66 & 17.05 \\
  rider      &  3.09 &  8.82 \\
  car        & 24.14 & 38.10 \\
  truck      &  6.02 & 10.66 \\
  bus        &  9.76 & 15.13 \\
  train      &  6.41 & 12.68 \\
  motorcycle &  3.62 & 10.66 \\
  bicycle    &  2.08 &  6.46 \\
  \hline
  mean       &  7.72 & 14.95 \\
  \hline
\end{tabular}
}
\end{table}

The results are around half of those from the validation set, demonstrating a high generalization error.
The test set not only presents more images but they were also acquired from different German cities.
The highest differences were from classes with small number of samples: truck, bus and train.
Therefore, the lower score in the test set is highly related to the dataset imbalance.

The test set results were obtained fine tuning all network layers. 
When fine tuning only the network last layers the highest mean AP obtained was 7.5.
This difference its likely caused by the classification prior mentioned in section \ref{sec:intro}.

When compared to DCME \cite{watanabe2018dcme} these results represent a considerable improvement, Table \ref{tab:test_dcme}.
Our scores are higher in all classes, specially classes with small number of samples (truck, bus, train and motorcycle).

\begin{table}[ht]
\centering
\caption{ DCME per class evaluation on Cityscapes. }
\label{tab:test_dcme}
\scalebox{1.1}{
\begin{tabular}{ l r r }
    \hline
    Class      &    AP & AP50\% \\
    \hline
    person     &  1.77 &  5.86 \\
    rider      &  0.71 &  3.33 \\
    car        & 15.53 & 25.65 \\
    truck      &  2.00 &  4.02 \\
    bus        &  4.30 &  8.30 \\
    train      &  4.57 &  9.98 \\
    motorcycle &  0.93 &  3.39 \\
    bicycle    &  0.33 &  1.35 \\
    \hline
    mean       &  3.77 &  7.73 \\
    \hline
\end{tabular}
}
\end{table}

Compared to other similar solution, ours is one third of Kendall \textit{et al.} \cite{kendall2018multi}, Table \ref{tab:test_multitask}.
However, they specialize the decoder in three subtasks, and since we repurpose the encoder our solution is more computational efficient.

\begin{table}[ht]
\centering
\caption{ Multitask Learning, per class evaluation on Cityscapes. }
\label{tab:test_multitask}
\scalebox{1.1}{
\begin{tabular}{ l r r }
    \hline
    Class      &    AP & AP50\% \\
    \hline
    person     & 19.22 & 38.10 \\
    rider      & 21.39 & 46.26 \\
    car        & 36.57 & 54.75 \\
    truck      & 18.80 & 28.44 \\
    bus        & 26.82 & 40.78 \\
    train      & 15.88 & 25.02 \\
    motorcycle & 19.39 & 42.21 \\
    bicycle    & 14.51 & 36.53 \\
    \hline
    mean       &  21.57 & 39.01 \\
    \hline
\end{tabular}
}
\end{table}

%% file: src/conclusion.tex
\section{Conclusion}
The proposed solution was able to solve the instance segmentation problem with a single CNN for image segmentation.
Differently from most of the approaches, this solution solves the partial occlusion problem.
Its computational cost is directly associated with the segmentation network and the input image resolution.
Compared to multi-task networks it presents a lower computational cost since the encoder is repurposed to solve the classification problem

%% file: src/acknowledgment.tex
\section*{Acknowledgment}

This research was funded by Sao Paulo Research Foundation (FAPESP) project: \#2015/26293-0.
This study was partially financed by Coordenacao de Aperfeicoamento de Pessoal de Nivel Superior (CAPES) - Finance Code 001.

%% file: root.bbl
\begin{thebibliography}{10}
\providecommand{\url}[1]{#1}
\csname url@samestyle\endcsname
\providecommand{\newblock}{\relax}
\providecommand{\bibinfo}[2]{#2}
\providecommand{\BIBentrySTDinterwordspacing}{\spaceskip=0pt\relax}
\providecommand{\BIBentryALTinterwordstretchfactor}{4}
\providecommand{\BIBentryALTinterwordspacing}{\spaceskip=\fontdimen2\font plus
\BIBentryALTinterwordstretchfactor\fontdimen3\font minus
  \fontdimen4\font\relax}
\providecommand{\BIBforeignlanguage}[2]{{%
\expandafter\ifx\csname l@#1\endcsname\relax
\typeout{** WARNING: IEEEtran.bst: No hyphenation pattern has been}%
\typeout{** loaded for the language `#1'. Using the pattern for}%
\typeout{** the default language instead.}%
\else
\language=\csname l@#1\endcsname
\fi
#2}}
\providecommand{\BIBdecl}{\relax}
\BIBdecl

\bibitem{watanabe2018dcme}
T.~Watanabe and D.~Wolf, ``Distance to center of mass encoding for instance
  segmentation,'' in \emph{The 21st IEEE International Conference on
  Intelligent Transportation Systems (ITSC)}, Hawaii, 2018, preprint at
  \url{https://arxiv.org/abs/1711.09060}.

\bibitem{garcia2017review}
A.~Garcia-Garcia, S.~Orts, S.~Oprea, V.~Villena-Martinez, and J.~G.
  Rodr{\'i}guez, ``A review on deep learning techniques applied to semantic
  segmentation,'' \emph{CoRR}, vol. abs/1704.06857, 2017, preprint at
  \url{https://arxiv.org/abs/1704.06857}.

\bibitem{chen2018encoder}
L.-C. Chen, Y.~Zhu, G.~Papandreou, F.~Schroff, and H.~Adam, ``Encoder-decoder
  with atrous separable convolution for semantic image segmentation,'' in
  \emph{European Conference in Computer Vision (ECCV)}, Munich, Germany,
  September 2018.

\bibitem{Cordts2016Cityscapes}
M.~Cordts, M.~Omran, S.~Ramos, T.~Rehfeld, M.~Enzweiler, R.~Benenson,
  U.~Franke, S.~Roth, and B.~Schiele, ``The cityscapes dataset for semantic
  urban scene understanding,'' in \emph{Proc. of the IEEE Conference on
  Computer Vision and Pattern Recognition (CVPR)}, 2016.

\bibitem{he2017mask}
K.~He, G.~Gkioxari, P.~Doll{\'a}r, and R.~Girshick, ``Mask r-cnn,'' in
  \emph{IEEE International Conference on Computer Vision (ICCV)}, 2017,
  preprint at \url{https://arxiv.org/abs/1703.06870}.

\bibitem{ren2015faster}
S.~Ren, K.~He, R.~Girshick, and J.~Sun, ``Faster r-cnn: Towards real-time
  object detection with region proposal networks,'' in \emph{Advances in neural
  information processing systems}, 2015, pp. 91--99.

\bibitem{liu2018path}
S.~Liu, L.~Qi, H.~Qin, J.~Shi, and J.~Jia, ``Path aggregation network for
  instance segmentation,'' in \emph{Proceedings of IEEE Conference on Computer
  Vision and Pattern Recognition (CVPR)}, 2018, preprint at
  \url{https://arxiv.org/abs/1803.01534}.

\bibitem{uhrig2016pixel}
J.~Uhrig, M.~Cordts, U.~Franke, and T.~Brox, ``Pixel-level encoding and depth
  layering for instance-level semantic labeling,'' in \emph{German Conference
  on Pattern Recognition}.\hskip 1em plus 0.5em minus 0.4em\relax Springer,
  2016, pp. 14--25.

\bibitem{de2017semantic}
B.~D. Brabandere, D.~Neven, and L.~V. Gool, ``Semantic instance segmentation
  for autonomous driving,'' in \emph{2017 IEEE Conference on Computer Vision
  and Pattern Recognition Workshops (CVPRW)}, July 2017, pp. 478--480.

\bibitem{bai2017deep}
M.~Bai and R.~Urtasun, ``Deep watershed transform for instance segmentation,''
  in \emph{2017 IEEE Conference on Computer Vision and Pattern Recognition
  (CVPR)}, July 2017, pp. 2858--2866.

\bibitem{liu2017sgn}
S.~Liu, J.~Jia, S.~Fidler, and R.~Urtasun, ``Sgn: Sequential grouping networks
  for instance segmentation,'' in \emph{The IEEE International Conference on
  Computer Vision (ICCV)}, Oct 2017.

\bibitem{kendall2018multi}
A.~Kendall, Y.~Gal, and R.~Cipolla, ``Multi-task learning using uncertainty to
  weigh losses for scene geometry and semantics,'' in \emph{Proceedings of the
  IEEE Conference on Computer Vision and Pattern Recognition (CVPR)}, 2018.

\bibitem{simonyan2014very}
K.~Simonyan and A.~Zisserman, ``Very deep convolutional networks for
  large-scale image recognition (vgg),'' \emph{arXiv preprint arXiv:1409.1556},
  2014, preprint at \url{https://arxiv.org/abs/1409.1556}.

\bibitem{he2016deep}
K.~He, X.~Zhang, S.~Ren, and J.~Sun, ``Deep residual learning for image
  recognition,'' in \emph{Proceedings of the IEEE Conference on Computer Vision
  and Pattern Recognition}, 2016, pp. 770--778.

\bibitem{jia2014caffe}
Y.~Jia, E.~Shelhamer, J.~Donahue, S.~Karayev, J.~Long, R.~Girshick,
  S.~Guadarrama, and T.~Darrell, ``Caffe: Convolutional architecture for fast
  feature embedding,'' in \emph{Proceedings of the 22nd ACM international
  conference on Multimedia}.\hskip 1em plus 0.5em minus 0.4em\relax ACM, 2014,
  pp. 675--678.

\bibitem{badrinarayanan2015segnet}
V.~Badrinarayanan, A.~Kendall, and R.~Cipolla, ``Segnet: A deep convolutional
  encoder-decoder architecture for image segmentation,'' in \emph{IEEE
  Transactions on Pattern Analysis and Machine Intelligence}.\hskip 1em plus
  0.5em minus 0.4em\relax IEEE, 2017.

\bibitem{hariharan2014simultaneous}
B.~Hariharan, P.~Arbel{\'a}ez, R.~Girshick, and J.~Malik, ``Simultaneous
  detection and segmentation,'' in \emph{European Conference on Computer
  Vision}.\hskip 1em plus 0.5em minus 0.4em\relax Springer, 2014, pp. 297--312,
  preprint at \url{https://arxiv.org/abs/1407.1808}.

\bibitem{lin2014microsoft}
T.-Y. Lin, M.~Maire, S.~Belongie, J.~Hays, P.~Perona, D.~Ramanan,
  P.~Doll{\'a}r, and C.~L. Zitnick, ``Microsoft coco: Common objects in
  context,'' in \emph{European conference on computer vision}.\hskip 1em plus
  0.5em minus 0.4em\relax Springer, 2014, pp. 740--755, preprint at
  \url{https://arxiv.org/abs/1405.0312}.

\end{thebibliography}
